\documentclass[9pt, conference]{IEEEtran}
\IEEEoverridecommandlockouts
\usepackage{cite}
\usepackage{amsmath,amssymb,amsfonts}
\usepackage{algorithmic}
\usepackage{graphicx}
\usepackage{textcomp}
\usepackage{CJKutf8}
\usepackage{CJK}
\usepackage{xcolor}
\usepackage{tabularray}
\usepackage{multirow}
\usepackage{booktabs} 
\usepackage{makecell}
\usepackage{rotating}
\usepackage{tabularray}
\usepackage{diagbox}
\usepackage{ulem}
\usepackage{utfsym}
\usepackage{fontawesome}
\usepackage{amsmath}

\def\BibTeX{{\rm B\kern-.05em{\sc i\kern-.025em b}\kern-.08em
    T\kern-.1667em\lower.7ex\hbox{E}\kern-.125emX}}
\begin{document}

\title{Investigating
Numerical Translation with Large Language Models}

\author{Wei Tang, Jiawei Yu, Yuang Li, Yanqing Zhao, Weidong Zhang, Wei Feng, Min Zhang, Hao Yang\\  Huawei Translation Services Center, China\\
\small{\{tangwei133, liyuang3, zhaoyanqing, zhangweidong17, fengwei31, zhangmin186, yanghao30\}@huawei.com}}

\maketitle

\begin{abstract}
The inaccurate translation of numbers can lead to significant security issues, ranging from financial setbacks to medical inaccuracies. While large language models (LLMs) have made significant advancements in machine translation, their capacity for translating numbers has not been thoroughly explored. This study focuses on evaluating the reliability of LLM-based machine translation systems when handling numerical data. In order to systematically test the numerical translation capabilities of currently open source LLMs, we have constructed a numerical translation dataset between Chinese and English based on real business data, encompassing ten types of numerical translation.  Experiments on the dataset indicate that errors in numerical translation are a common issue, with most open-source LLMs faltering when faced with our test scenarios. Especially when it comes to numerical types involving large units like ``million", ``billion", and \begin{CJK}{UTF8}{gbsn}``亿" \end{CJK}, even the latest llama3.1 8b model can have error rates as high as 20\%. Finally, we introduce three potential strategies to mitigate the numerical mistranslations for large units.
\end{abstract}


\section{Introduction}
Numerical translation is ubiquitous in our daily lives, especially in the fields of finance and technology, playing a crucial role. 
In comparison to translating plain text, numerical translation typically demands a high level of precision as it may involve mathematical operations (such as unit conversions) or other precise information.
As a result, such as a decimal point, an extra zero, or a missing zero, can have a significant impact on the alignment results. 
For instance, in the medical field, mistranslating the number of confirmed cases of a contagious disease such as COVID-19 can exacerbate public health misinformation. 
Consequently, numerical translation allows for little margin of error, in contrast to plain text translation, where a certain amount of ambiguity may be more permissible.

Another important distinction from plain text translation lies in the inadequacy of de facto evaluation metrics such as BLEU \cite{bleu} or COMET\cite{comet} for numerical translation. These metrics may fail to capture a minor numerical translation error that carries only a nominal penalty, often originating from a single-token mistranslation. Moreover, the diversity of data formats in numerical translation, encompassing Arabic numerals, Roman numerals, and more, further complicates the evaluation process.
For instance, in the context of Chinese-English translation, ``32.6\begin{CJK}{UTF8}{gbsn}万亿 \end{CJK}" could be translated as "32.6 trillion," "thirty-two trillion six hundred billion," or "32,600,000,000,000"; similarly, "\begin{CJK}{UTF8}{gbsn}第二\end{CJK}" might be translated as "second" or "2nd".

As numerical translation challenges constitute a minor segment of translation tasks, scholarly references on this domain are notably scarce. Only a handful of studies, such as \cite{123}, have endeavored to assess the effectiveness of Neural Machine Translation (NMT) models across four key categories of numerical translation: integers, decimals, numerals, and separators. It was found that both commercial translation systems and open-source NMT models do not guarantee 100\% accuracy in translating numerical data. In specific contexts (e.g., numerals) , the accuracy rate for numerical translations may drop below 60\%, a threshold considered unacceptable for critical financial applications. However, in \cite{123}, the testing examples are created using predefined prompts, potentially limiting the ability to accurately assess the numerical translation capabilities of current NMT models in real-world scenarios.
Furthermore,
in recent years, the rapid advancement of large language models (LLMs)\cite{LLM1,LLM2,LLM3} has highlighted their superior performance compared to traditional small neural models in various fundamental natural language processing (NLP) tasks. Consequently, many MT researchers\cite{li2024eliciting, mtllm} have turned to LLMs for their versatility and natural fluency in generation tasks. However, the extent to which LLMs excel in the field of numerical translation remains unexplored.  

Here we take large unit translation as an example and compare several prominent LLMs, such as the latest Llama3.1 series\cite{llama3}, the translation LLM Tower-7B\cite{tower_llm_2024}, and the Chinese LLM Qwen\cite{qwen2}. Besides, we also include the commercial translation tool Google Translate for comparison. As shown in Table~\ref{large unit}, apart from Google Translate, none of the LLMs can accurately translate this Chinese sentence that contains various large units. The reason for this inaccuracy is that LLMs struggle to accurately interpret the conversion relationships between units and execute precise mathematical computations required for translating this sentence correctly.

\begin{table*}[h]
\centering
\caption{Examples of numerical translation by LLMs on large units }

\renewcommand\arraystretch{1}
\label{large unit}
 \resizebox{1.0\linewidth}{!}{
\begin{tabular}{l}
  \toprule
  \textbf{Source}: \begin{CJK}{UTF8}{gbsn}某公司去年的年收入超过了\fbox{1000亿}美元，净利润达到\fbox{5000万}美元，总资产达到\fbox{三千五百亿}美元，其中包括\fbox{134亿}美元的现金储备。\end{CJK} \\
  \midrule
   \textbf{Google}: \\A company's revenue last year exceeded \$\textcolor{green}{100 billion}, reaching \$\textcolor{green}{50 million}, and total assets reached \$\textcolor{green}{350 billion}, including \$\textcolor{green}{13.4 billion} in cash reserves.\\
   \midrule
  \textbf{TowerInstruct-7B-v0.2}:\\
  A company's annual revenue last year exceeded \$\textcolor{green}{100 billion} , net profit reached  \$\textcolor{green}{50 million}, total assets reached  \$\textcolor{red}{35 billion}, including  \$\textcolor{red}{1.34 billion} in cash reserves.\\
   \midrule
  \textbf{Llama-3.1-8B}: \\A company's revenue last year exceeded \$\textcolor{red}{10 billion}, net profit reached \$\textcolor{green}{50 million}, and total assets reached \$\textcolor{red}{35 billion}, including \$\textcolor{red}{3.4 billion} in cash reserves.\\
  \midrule
  \textbf{Llama-3.1-70B}: \\Last year, a company's annual revenue exceeded \$\textcolor{green}{100 billion}, with a net profit of \$\textcolor{red}{500 million} and total assets of \$\textcolor{green}{350 billion}, including \$\textcolor{green}{13.4 billion} in cash reserves.\\
  \midrule
    \textbf{Llama-3.1-405B}:\\ A company's annual revenue exceeded \$\textcolor{green}{100 billion} last year, with a net profit of \$\textcolor{green}{50 million} and total assets of \$\textcolor{red}{3.5 trillion}, including \$\textcolor{green}{13.4 billion} in cash reserves.\\
    \midrule
  \textbf{Qwen-72B}:\\ A company's annual revenue exceeded \$\textcolor{green}{100 billion} last year, with a net profit of \$\textcolor{red}{500 million} and total assets of \$\textcolor{green}{350 billion}, including \$\textcolor{green}{13.4 billion} in cash reserves.\\
  \bottomrule
\end{tabular}}
\end{table*}

To deeply explore the numerical translation capacity of LLMs,  in this paper, we focus specifically on Chinese-English translation and construct bi-directional translation datasets between Chinese and English based on real business data. Furthermore, we summarize ten common numerical translation types, thoroughly investigating the performance of current LLMs in numerical translation. Considering the diversity of number formats,  we manually generate all possible translation results of the numerical parts in the original sentences as labels. This means that as long as a translation matches any of these labels, it will be considered a correct translation. Utilizing the proposed numerical translation dataset, we test popular translation LLMs and generic LLMs. Experimental results indicate that no LLMs can perform well in all numerical translation types and they perform worst in large unit translation overall. Finally, we introduced three strategies to enhance the numerical translation results of LLMs in the large unit type.

\begin{table*}[]
\centering
\caption{Statistics of number translation data of ten types between English and Chinese }
\renewcommand\arraystretch{1.1}
\label{dataset}
 \resizebox{1.0\linewidth}{!}{
\begin{tabular}{c|ccc|ccc}
\toprule
\multirow{2}{*}{\textbf{Type}}                                               & \multicolumn{3}{c}{\textbf{EN-ZH}}   & \multicolumn{3}{c}{\textbf{ZH-EN}}   \\ \cline{2-7}    & \textbf{Count} & \textbf{Example} & \textbf{Reference} & \textbf{Count} & \textbf{Example} & \textbf{Reference} \\ \hline
\begin{tabular}[c]{@{}c@{}}Large unit\end{tabular}         &    693   &   2.82 billion      &  \makecell[c]{[2820000000, 2,820,000,000, 28.2\begin{CJK}{UTF8}{gbsn}亿\end{CJK}, \begin{CJK}{UTF8}{gbsn}二十八点二亿\end{CJK}\\\begin{CJK}{UTF8}{gbsn}28亿2千万\end{CJK}, \begin{CJK}{UTF8}{gbsn} 二十八亿两千万\end{CJK}, \begin{CJK}{UTF8}{gbsn} 二十八亿二千万 \end{CJK}]}        &   197   & \begin{CJK}{UTF8}{gbsn} 1.43亿\end{CJK}        &  \makecell[c]{[143000000, 143,000,000,\\143 million, 0.143 billion] }        \\ 
\begin{tabular}[c]{@{}c@{}}Range\end{tabular}                &    165   &     between 300 and 500    &    [300\textasciitilde500, 300-500, 300\begin{CJK}{UTF8}{gbsn}到\end{CJK}500, 300\begin{CJK}{UTF8}{gbsn}至\end{CJK}500]       &   413    &  10\textasciitilde1440       &  [10-1440, 10\textasciitilde1440, 10 to 1440]         \\ 
\begin{tabular}[c]{@{}c@{}}Decimal\end{tabular}              &   252    &     1.85    &    [ 2.5,   \begin{CJK}{UTF8}{gbsn}二点五\end{CJK}]     &    252  &   3.525      &    [3.525]       \\ 
\begin{tabular}[c]{@{}c@{}}Number String\end{tabular}       &    190   &     01074316-002    &     01074316-002      &    1174   &    00326264        &  [00326264]         \\ 
\begin{tabular}[c]{@{}c@{}}Fraction\end{tabular}             &   45    &    half     &    [$\frac{1}{2}$, 1/2, \begin{CJK}{UTF8}{gbsn}二分之一\end{CJK}, \begin{CJK}{UTF8}{gbsn}2分之1\end{CJK},\begin{CJK}{UTF8}{gbsn}一半\end{CJK}]       &   23    &    \begin{CJK}{UTF8}{gbsn}四分之一\end{CJK}     &  \makecell[c]{[1/4, one-fourth, one fourth, \\one in four, 1 in 4, quarter, $\frac{1}{4}$]}        \\ 
\begin{tabular}[c]{@{}c@{}}Ratio\end{tabular}                &   36    &    1:1     &    [1:1,  \begin{CJK}{UTF8}{gbsn}1比1\end{CJK}, \begin{CJK}{UTF8}{gbsn}一比一\end{CJK} ]       &    51   &    16:9     &    [16:9]       \\ 
\begin{tabular}[c]{@{}c@{}}Negative Numbers\end{tabular}     &    83   &    minus two     &    [-2, \begin{CJK}{UTF8}{gbsn}负2\end{CJK},  \begin{CJK}{UTF8}{gbsn}负二\end{CJK} ]       &    41   &    -105     &   [-105, minus 105, minus one hundred and five]        \\ 
\begin{tabular}[c]{@{}c@{}}Formula\end{tabular}              &    19   &   48 x 48      & [48*48, 48x48, 48 * 48]          &  32     &    1+1     &  [1+1, 1 + 1, one plus one]         \\ 
\begin{tabular}[c]{@{}c@{}}Ordinal\end{tabular}             &    336   &       62nd  &    [\begin{CJK}{UTF8}{gbsn}第62\end{CJK}, \begin{CJK}{UTF8}{gbsn}第六十二\end{CJK}]       &    700   &   \begin{CJK}{UTF8}{gbsn}第十一\end{CJK}      & [eleventh, 11th]          \\ 
\begin{tabular}[c]{@{}c@{}}Special\end{tabular}              &   131    &     three-fold    &   [\begin{CJK}{UTF8}{gbsn}3倍\end{CJK}, \begin{CJK}{UTF8}{gbsn}三倍\end{CJK}]        &    97   &   \begin{CJK}{UTF8}{gbsn}700万像素\end{CJK}     &    \makecell[c]{[7.0 MP, 7.0 megapixel,\\ 7.0-megapixel, 7 megapixels]}      \\ 
\bottomrule
\end{tabular}
}

\footnotesize{Count denotes the number of the specific numerical translation type. It is worth noting that there is a significant difference in the number of types translated for each number. This is because the data is collected from real business data, conforming to real data distributions.}
\end{table*}

\begin{table*}
\centering
\caption{Main results of different LLMs on numerical translation on numerical translation dataset .}
\renewcommand\arraystretch{1.23}
\label{tb:main results1}
\resizebox{1\textwidth}{!}{
\begin{tabular}{c|c|c|cccccccccc|c} 
\toprule
                                            & \multicolumn{2}{c|}{Models}                                 & \textbf{Large unit} & \textbf{Range} & \textbf{Decimal} & \begin{tabular}[c]{@{}c@{}}\textbf{Number}\\\textbf{~String}\end{tabular} & \textbf{Fraction} & \textbf{Ratio} & \begin{tabular}[c]{@{}c@{}}\textbf{Negative }\\\textbf{Number}\end{tabular} & \textbf{Formula} & \textbf{Ordinal} & \textbf{Special} & \textbf{Avg.}   \\ 
\hline
\multirow{10}{*}{\rotatebox{90}{\textbf{EN-ZH }}} & \multirow{3}{*}{\rotatebox{90}{\textbf{T-LLMs}}}  & NLLB-200-3.3B & 0.844               & 0.896          & 0.968            & 0.926                                                                     & 0.933             & 0.916          & 0.915                                                                       & 0.684            & 0.928            & 0.908            & \textbf{0.891}  \\
                                            &                                             & BLOOMZ-7B-MT & 0.607               & 0.903          & 0.968            & 0.952                                                                     & 0.733             & 0.833          & 0.903                                                                       & 0.631            & 0.922            & 0.847            & \textbf{0.829}  \\
                                            &                                             & Tower-7B      & 0.965               & 0.830          & 0.992            & 0.968                                                                     & 1.00              & 0.833          & 0.915                                                                       & 0.894            & 0.970            & 0.916            & \textbf{0.928}  \\ 
\cline{2-14}
                                            & \multirow{6}{*}{\rotatebox{90}{\textbf{G-LLMs}}}  & Mistral-7B    & 0.445               & 0.884          & 0.992            & 0.978                                                                     & 0.822             & 0.805          & 0.939                                                                       & 0.894            & 0.863            & 0.862            & \textbf{0.848}  \\
                                            &                                             & Llama3.1-8B   & 0.743               & 0.939          & 0.996            & 0.973                                                                     & 0.888             & 0.972          & 0.987                                                                       & 0.947            & 0.967            & 0.931            & \textbf{0.934}  \\
                                            &                                             & Llama3.1-70B  & 0.904               & 0.933          & 0.992            & 0.978                                                                     & 0.955             & 0.833          & 0.975                                                                       & 0.842            & 0.970            & 0.969            & \textbf{0.935}  \\
                                            &                                             & Qwen2-7B      & 0.943               & 0.921          & 0.992            & 0.984                                                                     & 0.977             & 0.916          & 0.939                                                                       & 0.894            & 0.973            & 0.977            & \textbf{0.951}  \\
                                            &                                             & Qwen2-70B     & 0.974               & 0.927          & 1.00              & 0.989                                                                     & 0.977             & 0.888          & 0.975                                                                       & 0.894            & 0.988            & 0.977            & \textbf{0.958}  \\
                                            &                                             & GLM-4-9B-Chat & 0.949               & 0.951          & 0.992            & 0.984                                                                     & 1.00              & 0.972          & 0.951                                                                       & 0.947            & 0.985            & 0.992            & \textbf{0.972}  \\ 
\cline{2-14}
                                            & \multicolumn{2}{c|}{\textbf{ Avg.}}                         & \textbf{0.819}      & \textbf{0.909} & \textbf{0.988}   & \textbf{0.970}                                                            & \textbf{0.920}    & \textbf{0.885} & \textbf{0.944}                                                              & \textbf{0.847}   & \textbf{0.951}   & \textbf{0.931}   & \diagbox{}{}    \\ 
\midrule
\multirow{10}{*}{\rotatebox{90}{\textbf{ZH-EN}}}  & \multirow{3}{*}{\rotatebox{90}{\textbf{T-LLMs }}} & NLLB-200-3.3B & 0.908               & 0.736          & 0.996            & 0.959                                                                     & 0.869             & 0.960          & 0.731                                                                       & 0.937            & 0.958            & 0.814            & \textbf{0.886}  \\
                                            &                                             & BLOOMZ-7B-MT & 0.111               & 0.830          & 0.980            & 0.965                                                                     & 0.652             & 0.843          & 0.926                                                                       & 0.656            & 0.912            & 0.773            & \textbf{0.764}  \\
                                            &                                             & Tower-7B      & 0.949               & 0.946          & 0.988            & 0.993                                                                     & 0.826             & 1.00           & 1.00                                                                        & 0.968            & 0.972            & 0.907            & \textbf{0.954}  \\ 
\cline{2-14}
                                            & \multirow{6}{*}{\rotatebox{90}{\textbf{T-LLMs }}} & Mistral-7B    & 0.578               & 0.830          & 0.996            & 0.985                                                                     & 0.782             & 1.00           & 0.951                                                                       & 0.625            & 0.975            & 0.783            & \textbf{0.850}  \\
                                            &                                             & Llama3.1-8B   & 0.873               & 0.949          & 1.00             & 0.995                                                                     & 0.782             & 1.00           & 0.975                                                                       & 0.937            & 0.952            & 0.896            & \textbf{0.935}  \\
                                            &                                             & Llama3.1-70B  & 0.827               & 0.912          & 0.984            & 0.987                                                                     & 0.913             & 0.941          & 0.926                                                                       & 0.843            & 0.987            & 0.948            & \textbf{0.926}  \\
                                            &                                             & Qwen2-7B      & 0.822               & 0.924          & 0.996            & 0.998                                                                     & 0.782             & 0.980          & 0.951                                                                       & 0.875            & 0.958            & 0.896            & \textbf{0.918}  \\
                                            &                                             & Qwen2-70B     & 0.969               & 0.944          & 0.992            & 0.992                                                                     & 0.826             & 1.00           & 0.975                                                                       & 0.875            & 0.965            & 0.907            & \textbf{0.944}  \\
                                            &                                             & GLM-4-9B-Chat & 0.989               & 0.976          & 1.00             & 1.00                                                                      & 0.869             & 1.00           & 1.00                                                                        & 0.906            & 0.971            & 0.917            & \textbf{0.962}  \\ 
\cline{2-14}
                                            & \multicolumn{2}{c|}{\textbf{Avg. }}                         & \textbf{0.780}      & \textbf{0.894} & \textbf{0.992}   & \textbf{0.986}                                                            & \textbf{0.811}    & \textbf{0.969} & \textbf{0.937}                                                              & \textbf{0.846}   & \textbf{0.961}   & \textbf{0.871}   & \diagbox{}{}    \\
\bottomrule
\end{tabular}
}
\end{table*}

\section{The proposed numerical translation dataset}
In this section, we elaborate on the ten types of numerical translation included in the Chinese-English and English-Chinese translation datasets proposed. Detailed data can be found in Table~\ref{dataset}.
\begin{itemize}
    \item \textbf{Large unit,} which involves converting numerical values from one system of measurement to another, such as converting ``\begin{CJK}{UTF8}{gbsn}亿\end{CJK}" as ``million" or ``billion". This type of translation is crucial for accurate communication of measurements and quantities in different units.
\item \textbf{Range,} which deals with converting numerical ranges or intervals from one language to another. This can include translating ranges of values like ``1-10" or ``[10000, 1000000]" into the respective target language while preserving the correct numerical format and meaning.
\item \textbf{Decimal,} which focuses on accurately translating decimal numbers between languages. This includes handling both whole and fractional parts of numbers, ensuring that decimal points and decimal separators are correctly represented in the target language.

\item \textbf{Number String,} which is very common in everyday life,, such as employee ID numbers, phone numbers, and patent numbers, among others. We aim for these number strings to remain unchanged after translation. However, LLMs may mistakenly identify them as numerical values, resulting in translations with a numeric format. For example, 
``Tom's employee ID is 10000000" might be incorrectly translated as ``Tom's employee ID is 10,000,000".

\item \textbf{Fraction,} whose difficulty lies in the fact that, whether in Chinese or English, there are diverse ways to express fractions, which can be confusing for models. This includes converting fractions like 1/2,  $\frac{1}{2}$, \begin{CJK}{UTF8}{gbsn}二分之一\end{CJK} , or one fourth into the equivalent format in the target language while maintaining the correct fractional representation.

\item \textbf{Ratio,} which involves converting ratios and proportions from one language to another, ensuring that ratios like 2:1 or 3:2 are correctly interpreted and expressed in the target language to convey the intended meaning accurately.
\item \textbf{Negative number,}  which focuses on accurately translating negative numerical values between languages. This includes handling negative integers and negative decimal numbers to ensure that the negative sign is correctly represented in the target language.
\item \textbf{Formula,} which involves converting mathematical formulas and expressions from one language to another. This type of translation is essential for accurately communicating mathematical relationships, equations, and calculations in the target language while preserving their original meaning.
\item \textbf{Ordinal,} which deals with converting ordinal numbers (e.g., first, second, 1st, 2nd) from one language to another. This type of translation ensures that the correct ordinal indicators are used to represent the order or rank of items or entities in the target language.
\item \textbf{Special,} which covers specific numerical formats that do not fall into the other defined types. This type of translation may include handling special symbols, notations, or formats that require specific treatment to accurately convey the numerical information in the target language.
\end{itemize}

\section{Enhancing Performance: Evaluation of Current LLMs and potential strategies}
In this section, we demonstrate the performance of current LLMs on the proposed numerical translation dataset. 
The LLMs we tested are mainly divided into generic LLMs (G-LLMs) and translation LLMs (T-LLMs), where the G-LLMs include Llama3.1-8B \cite{llama3}, Llama3.1-70B \cite{llama3}, 
Qwen2-7B-Instruct \cite{qwen2}, 
Qwen2-72B-Instruct \cite{qwen2}, GLM-4-9B-Chat and 
Mistral-7B \cite{jiang2023mistral7b}, while the T-LLMs consist of TowerInstruct-7B \cite{tower_llm_2024}, BLOOMZ-7B-MT \cite{muennighoff2022crosslingual} and NLLB-200-3.3B \cite{nllb}. 
For better evaluation, we use Pass Rate (PR) \cite{123} as the evaluation metric, which is the proportion of predictions where the system translates the numerical component perfectly, i.e., as long as the model's translation result matches any one of the results in the reference list, it is considered a pass.
\subsection{Error Analysis}
As depicted in Table~\ref{tb:main results1}, T-LLMs and G-LLMs exhibit distinct advantages in various numerical translation scenarios, yet no single model emerges as superior across all types. Notably, GLM-4-9b-chat emerges as the top performer from a model-centric standpoint, boasting an average accuracy of 97.2\% in EN-ZH and 96.2\% in ZH-EN, with consistent accuracy exceeding 94\% across all numerical types. However, LLMs consistently struggle with translations involving large units, registering averages of 81.9\% and 78\% in EN-ZH and ZH-EN, respectively, underscoring their vulnerability in unit conversion.
Regarding model size, it is evident that larger models do not invariably guarantee enhanced numerical translation precision. A case in point is the comparison between Llama3.1-8b and Llama3.1-70B, where the latter notably enhances accuracy in large units and fractions while experiencing a pronounced decline in accuracy for ratios and formulas.

Additionally, we found that LLMs often produce overtranslations or misunderstand meanings, such as translating ``ranked 4th place" as ``\begin{CJK}{UTF8}{gbsn}获得铜牌\end{CJK} (won bronze medal)," or translating ``five out of six unsafe signals" as ``\begin{CJK}{UTF8}{gbsn}五分之六的危险信号\end{CJK} (six out of five unsafe signals)" or ``\begin{CJK}{UTF8}{gbsn}六个危险信号中有五个\end{CJK}$\cdots$ (there are five $\cdots$ from six unsafe signals)." These results indicate that numerical translation remains challenging for existing LLMs, requiring further exploration and improvement.

We argue that LLMs' architecture, such as the Transformers model, is better at handling sequential data and language patterns rather than numerical computations. Although these models can generate translation results by learning numerical patterns in the text, they do not possess the ability to perform unit conversions. Moreover, even in numerical operations, LLMs face precision issues in floating-point calculations. Due to limitations in underlying hardware and algorithms, small errors often arise. These errors accumulate gradually in large-scale calculations, thereby affecting the accuracy of the final results.

\begin{table*}
\centering
\caption{Prompts of ICL and COT strategies and Case study.  }
\renewcommand\arraystretch{1.3}
\label{case}
\begin{tblr}{
  cell{1}{1} = {c=3}{},
  cell{2}{1} = {c},
  cell{2}{2} = {c},
  cell{2}{3} = {c},
  cell{3}{1} = {c},
  cell{4}{1} = {c},
  cell{5}{1} = {c},
  vline{2-3} = {2-4}{},
  hline{1-5} = {1-3}{},
}
{\textbf{\textbf{Source:~}}
\begin{CJK}{UTF8}{gbsn}某公司去年的年收入超过了1000亿美元，净利润达到5000万美元，总资产达到三千五百亿美元，其中包括134亿美元的现金储备。\end{CJK}
\\\textbf{\textbf{Llama3.1-8B:}}\textcolor[rgb]{0.416,0.671,0.451}{\textbf{\textbf{~}}}\textcolor[rgb]{0.416,0.671,0.451}{}A company’s revenue last year exceeded \$\textcolor{red}{10 billion}, net profit reached \$50 million, and total assets\\reached \$\textcolor{red}{35 billion}, including \$\textcolor{red}{3.4 billion}~in cash reserves.} &                                                                                                                                                                                                                                                                          &                                                                                                                                                                                                                                                                                                                                                                                                                                                                                                                                                                                                                                                                                                                                      &  \\
\textbf{\textbf{Strategy}}                                                                                                                                                                                                                                                                                                                                                                                                                       & \textbf{\textbf{Prompt}}                                                                                                                                                                                                                                                 & \textbf{Output}                                                                                                                                                                                                                                                                                                                                                                                                                                                                                                                                                                                                                                                                                                                      &  \\
\textbf{\textbf{ICL}}                                                                                                                                                                                                                                                                                                                                                                                                                            & {You are a good translator. Help me\\~translate the [Source] sentence into\\~[Target] sentence based on the given \\~unit conversion principle.\\Unit Convesion Principle:\\1 million = 100 \begin{CJK}{UTF8}{gbsn}万\end{CJK}\\1 billion = 10 \begin{CJK}{UTF8}{gbsn}亿\end{CJK}\\$\cdots$ }   & {~Here is the translation of the Chinese sentence into English:~\uline{\underline{A company's annual} }\\\uline{\underline{revenue last year exceeded \$\textcolor{red}{1 trillion}, net profit reached \$50 million, total assets} }\\\uline{\underline{reached \$\textcolor[rgb]{0.416,0.671,0.451}{350 billion},~ including \$\textcolor{red}{34 billion}~in cash reserves.}}~ ~Here's the breakdown\\~of the numerical parts: 1000\begin{CJK}{UTF8}{gbsn}亿\end{CJK}~~= \$\textcolor{red}{1 trillion}~(1,000,000,000,000), 5000\begin{CJK}{UTF8}{gbsn}万\end{CJK} = \$50\\million (\textcolor{red}{5,000,000}), 3500\begin{CJK}{UTF8}{gbsn}亿\end{CJK} = \$350 billion}                                                                                                                                                                                               &  \\
\textbf{COT}                                                                                                                                                                                                                                                                                                                                                                                                                                     & {You are a good translator. Help me \\translate the [Source] sentence into\\~[Target] sentence step by step.\\Please pay attention to the unit \\conversion between Chinese and\\~English and first translate the~\\numerical parts, and then translate \\the sentence.} & {Let's break down the Chinese sentence into numerical parts and then translate the \\sentence. **Numerical parts:**\\1. 1000\begin{CJK}{UTF8}{gbsn}亿美元\end{CJK} = 100,000,000,000 USD (\textcolor{red}{1 billion} USD)\\2. 5000\begin{CJK}{UTF8}{gbsn}万美元\end{CJK}= 50,000,000 USD (\textcolor{red}{5 million} USD)\\3. \begin{CJK}{UTF8}{gbsn}三千五百亿美元\end{CJK} = \textcolor{red}{35,500,000,000 }USD (\textcolor{red}{35.5} billion USD)\\4. 134\begin{CJK}{UTF8}{gbsn}亿美元\end{CJK} = 13,400,000,000 USD (13.4 billion USD)\\ **Translated sentence:**\\\uline{\underline{A company's annual revenue last year exceeded \textcolor[rgb]{0.416,0.671,0.451}{100 billion} USD, net profit} }\\\uline{\underline{reached 50 million USD, total assets reached \textcolor{red}{35.5 billion} USD, of which \textcolor[rgb]{0.416,0.671,0.451}{13.4 billion }}}\\\uline{\underline{USD was cash reserves.}}}            
\end{tblr}
\\
\footnotesize{Red denotes incorrect translations or inferences, while green indicates where the strategy is implemented to correct errors.}
\end{table*}

\subsection{Potential Strategies for improving large unit conversion }
\begin{figure}[!t]
	\centering
	\includegraphics[width=0.4\textwidth]{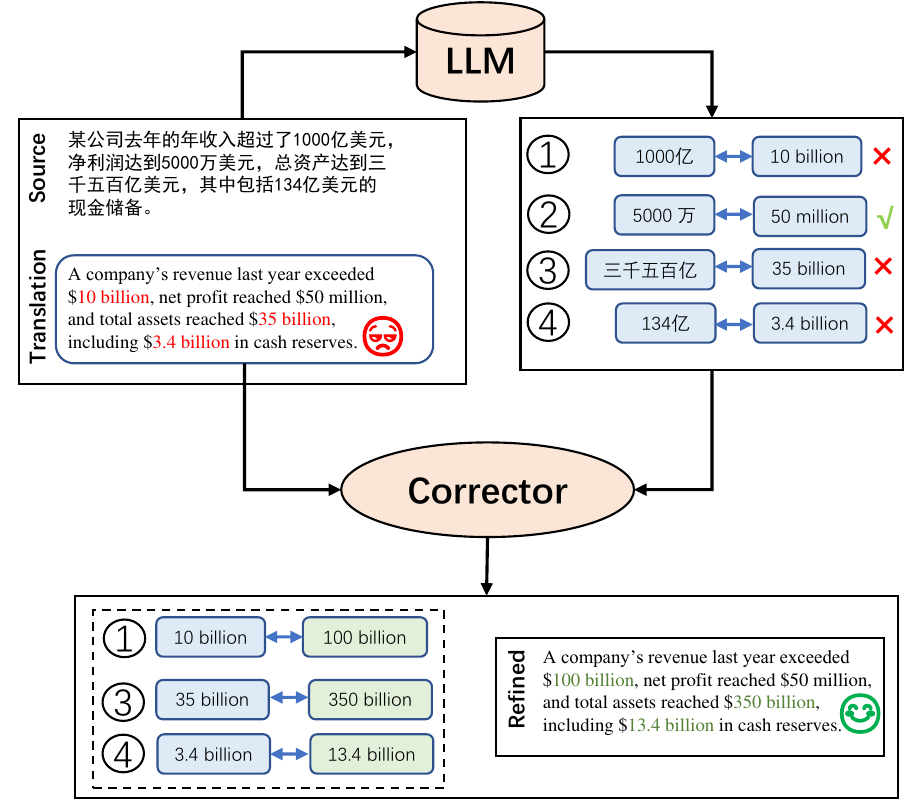} 
	\caption{The framework of PE strategy.}
	\label{model}
\end{figure}

Since LLMs struggle with translating large units, especially considering their prevalence among various numerical translation types, we present three commonly employed strategies aimed at enhancing numerical translation:
 (1) \textbf{In-context Learning (ICL)} \cite{DBLP:conf/acl/AgrawalZLZG23, zhu-etal-2024-towards-robust, ji-etal-2024-submodular}, which prompts LLMs with the specific unit conversion principles, aiming to make LLMs translate the large units based on the principles. (2) \textbf{Chain-of-Thought (COT)} \cite{peng-etal-2023-towards, DBLP:journals/corr/abs-2401-07037, DBLP:conf/acl/Chen0ZC0C24}, which prompt LLMs to composite the translation task and obtain the result step by step. (3) \textbf{Post-editing (PE)} \cite{brockmann-etal-2022-error, DBLP:conf/naacl/KiC24, DBLP:conf/naacl/KoneruEHN24}, which involves extracting numerical translation pairs (e.g., [``1 million", ``100 \begin{CJK}{UTF8}{gbsn}万\end{CJK}]) from the translation pairs. Here, we use LLM to finish the extraction task and prompts is designed as follows: \textit{
You are an excellent extractor of numerical translation pairs. Please extract all the numerical translation pairs from the given [Source]-[Target] translation pairs. Please output the extracted numerical translation pairs in the form of list without giving any explanation. Here is an example: [Source]: It will provide EUR 72.2 billion over 7 years in funding. [Target]: \begin{CJK}{UTF8}{gbsn}它将在7年内提供722亿欧元的资金。\end{CJK}
output:[(``72.2 billion", ``722\begin{CJK}{UTF8}{gbsn}亿\end{CJK}")]. Here is the [Source]-[Target] translation pair you need to extract:  [Source sentence],  [Target sentence] }.
Having extracted the numerical pairs, we use Python packages cn2an\footnote{https://github.com/Ailln/Cn2An.jl} and en2an\footnote{https://github.com/kdwycz/en2an} to convert them into digits (Arabic or local scripts), thus 
detecting the potential wrong translations. Subsequently, we replace the erroneous numerical translation with the correct digits format\footnote{Here, one can also transform the digit form into large unit form with a simple unit conversion function.}.
 The PE strategy use the Llama3.1-8B as the base model and its 
framework can be seen in the Figure~\ref{model}, and the prompts of ICL and COT are listed in Table~\ref{case}.


\subsection{Main Results}
To evaluate the three introduced strategies, we selected BLOOMZ-7B-MT, Mistral-7B, and Llama3.1-8B as the base models due to their limited translation capacity in the context of \textit{Large unit}. As depicted in Table~\ref{result}, the PE strategy consistently outperforms the ICL and COT strategies on both the EN-ZH and ZH-EN datasets across all base models in terms of the PR metric. This trend can be attributed to the fact that the ICL and COT strategies aim to enhance LLM translation accuracy during the inference process, yet they are unable to address the inherent challenge of LLMs in precise numerical calculations.
Table~\ref{case} presents a translation example where the base model Llama3.1-8B exhibits unit conversion errors (1000 \begin{CJK}{UTF8}{gbsn}亿\end{CJK} $\rightarrow$ 10 billion, \begin{CJK}{UTF8}{gbsn}三千五百亿\end{CJK} $\rightarrow$ 35 billion) and illusion issues (134 \begin{CJK}{UTF8}{gbsn}亿\end{CJK} $\rightarrow$ 3.4 billion). By using ICL and COT strategies, it is observed in their final translation outputs that ICL only corrects one translation error, while COT corrects two. Furthermore, we also found that some translations were incorrect in the reasoning process, but turned out to be correct in the final results. This highlights the limitation of LLMs in comprehending numerical calculations at their core.
In comparison to ICL and COT, the PE strategy focuses solely on extracting numerical translation pairs from LLMs, a task within their capabilities, and achieves superior performance by post-editing (detecting and correcting) translation results.

\begin{table}
\centering
\renewcommand\arraystretch{1.1}
\caption{Performance of different strategies on type of large unit.}
\label{result}
\resizebox{0.5\textwidth}{!}{
\begin{tabular}{c|cccccc} 
\toprule
\multirow{2}{*}{Strategies} & \multicolumn{2}{c|}{\textbf{BLOOMZ-7B-MT~}}                               & \multicolumn{2}{c|}{\textbf{Mistral-7B }}                                 & \multicolumn{2}{c}{\textbf{Llama3.1-8B}}  \\ 
\cline{2-7}
                            & \multicolumn{1}{c|}{\textbf{EN-ZH}} & \multicolumn{1}{c|}{\textbf{ZH-EN}} & \multicolumn{1}{c|}{\textbf{EN-ZH}} & \multicolumn{1}{c|}{\textbf{ZH-EN}} & \textbf{EN-ZH} & \textbf{ZH-EN}           \\ 
\hline
\textbf{Base}               & 0.607                               & 0.111                               & 0.445                               & 0.578                               & 0.743          & 0.873                    \\
\textbf{ICL}                & 0.641                               & 0.463                               & 0.489                               & 0.601                               & 0.768          & 0.897                    \\
\textbf{COT}                & 0.668                               & 0.451                               & 0.507                               & 0.597                               & 0.807          & 0.908                    \\
\textbf{PE}                 & \textbf{0.948}                               & \textbf{0.938}                               & \textbf{0.932}                               & \textbf{0.957}                               & \textbf{0.956}          & \textbf{0.963}                    \\
\bottomrule
\end{tabular}
}
\end{table}


\section{Conclusion}
In this paper, we summarize ten typical types of numerical translation and create a dataset for English-Chinese and Chinese-English numerical translation. We also investigate the performance of generic LLMs and translation LLMs on this dataset. The experiments show that the large models' performance on numerical translation is not satisfactory, especially in the translation of large numerical units, fractions, and other types that require numerical calculations. Additionally, we introduce and discuss three strategies for translating large numerical units and find that post-processing strategies can achieve relatively good results. We hope that future research will follow up on LLM-based numerical translation and propose more effective strategies.

\clearpage
\bibliographystyle{IEEEbib}
\bibliography{refs}

\end{document}